\documentclass[runningheads]{llncs}
\usepackage{graphicx}
\usepackage{amsmath}
\usepackage{algorithm}
\usepackage[noend]{algpseudocode}
\usepackage{amsmath,amssymb} 
\usepackage{color}

\begin{document}

\title{Reading Industrial Inspection Sheets by Inferring Visual Relations} 
\titlerunning{Reading Industrial Inspection Sheets} 


\author{Rohit Rahul\inst{1} \and
Arindam Chowdhury\inst{1} \and
Animesh\inst{1} \and
Samarth Mittal\inst{2} \and
Lovekesh Vig\inst{1}}
%

\authorrunning{R. Rahul et al.} 


\institute{ TCS Research, Delhi, India \and
BITS Pilani, Goa Campus, India\\
\email{rohit.rahul@tcs.com}}

\maketitle
\graphicspath{{./input_images/}}

\begin{abstract}
The traditional mode of recording faults in heavy factory equipment has been via handmarked inspection sheets, wherein a machine engineer manually marks the faulty machine regions on a paper outline of the machine. Over the years, millions of such inspection sheets have been recorded and the data within these sheets has remained inaccessible. However, with industries going digital and waking up to the potential value of fault data for machine health monitoring, there is an increased impetus towards digitization of these handmarked inspection records. To target this digitization, we propose a novel visual pipeline combining state of the art deep learning models, with domain knowledge and low level vision techniques, followed by inference of visual relationships. Our framework is robust to the presence of both static and non-static background in the document, variability in the machine template diagrams, unstructured shape of graphical objects to be identified and variability in the strokes of handwritten text. The proposed pipeline incorporates a capsule and spatial transformer network based classifier for accurate text reading, and a customized CTPN \cite{tian2016detecting} network for text detection in addition to hybrid techniques for arrow detection and dialogue cloud removal. We have tested our approach on a real world dataset of $50$ inspection sheets for large containers and boilers. The results are visually appealing and the pipeline achieved an accuracy of \textbf{87.1\%} for text detection and \textbf{94.6\%} for text reading.

\end{abstract}

\section{Introduction}
Industrial inspection of factory equipment is a common process in factory settings, involving inspection engineers conducting a physical examination of the equipment and subsequently marking faults on paper based inspection sheets. While many industries have digitized the inspection process \cite{agin1980computer}\cite{golnabi2007design}\cite{ar_inspection}, paper based inspection is still widely practiced, frequently followed by a digital scanning process. These paper based scans have data pertaining to millions of faults detected over several decades of inspections. Given the tremendous value of fault data for predictive maintainence, industries are keen to tap into the vast reservoir of fault data stored in the form of highly unstructured scanned inspection sheets and generate structured reports from them.  

However, there are several challenges associated with digitizing these reports ranging from image preprocessing and layout analysis to word and graphic item recognition \cite{marinai2008introduction}. There has been plenty of work in document digitization in general but very little prior work on digitization of inspection documents. In this paper, we have addressed the problem of information extraction from boiler and container inspection documents. The target document, as shown in Figure~\ref{fig:flow}, has multiple types of printed machine line diagrams, where each diagram is split into multiple zones corresponding to different components of the machine. The inspection engineer marks handwritten damage codes and comments against each component of the machine (machine zone). These comments are connected via a line or an arrow to a particular zone. Thus, the arrow acts as a connector that establishes the relationship between a text cloud containing fault codes, and a machine zone.

\section{Problem Description}

\begin{figure}
 \centering
 \includegraphics[scale={0.5}]{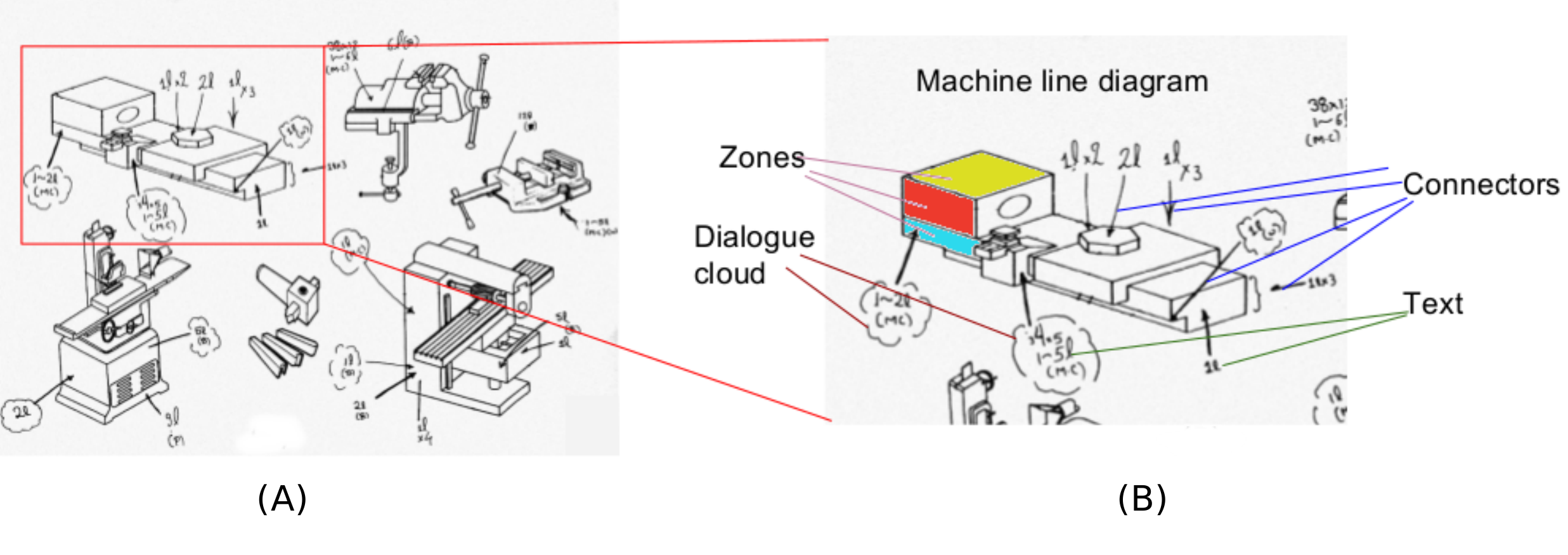}
 \caption{ (A) Overview of an inspection sheet (B) Essential components of the sheet.}
 \label{fig:flow}
\end{figure}

\noindent In this work, we strive to extract relevant information from industrial inspection sheets which contain multiple 3D orthogonal machine diagrams. Figure \ref{fig:flow}(A) shows one such inspection sheet consisting of $7$ machine diagrams. We define a \textit{set} as a collection of inspection sheets which contain identical machine diagrams while the individual machine diagrams in an inspection sheet are called \textit{templates}, as shown in Figure \ref{fig:temp}. Each template consists of multiple zones. In Figure \ref{fig:flow}(B) we mark individual zones with different colors. In industrial setting, an inspector goes around examining each machine. If he detects any damage in a machine then he identifies the zone where the damage has occured. He then draws an entity which we call as \textit{connector} as shown in Figure \ref{fig:flow}(B) and writes a damage code at the tail of the connector. Each code corresponds to one of the predefined damages that could occur in the machine. This damage code is shown as \text{text} in Figure \ref{fig:flow}(B). Often, the text is enclosed in a bubble or a cloud structure that carry no additional information of relevance but adds to the overall complexity of the problem. Our task is to localize and read the damage codes that are written on the inspection sheet and associate each damage code with the zone against which it is marked and store the information in a digital document. This allows firms to analyze data on their machines, that was collected over the years, with minimum efforts.

\section{Proposed Method}\label{sec:propMet}

\begin{figure}[!h]
\includegraphics[scale={0.06}]{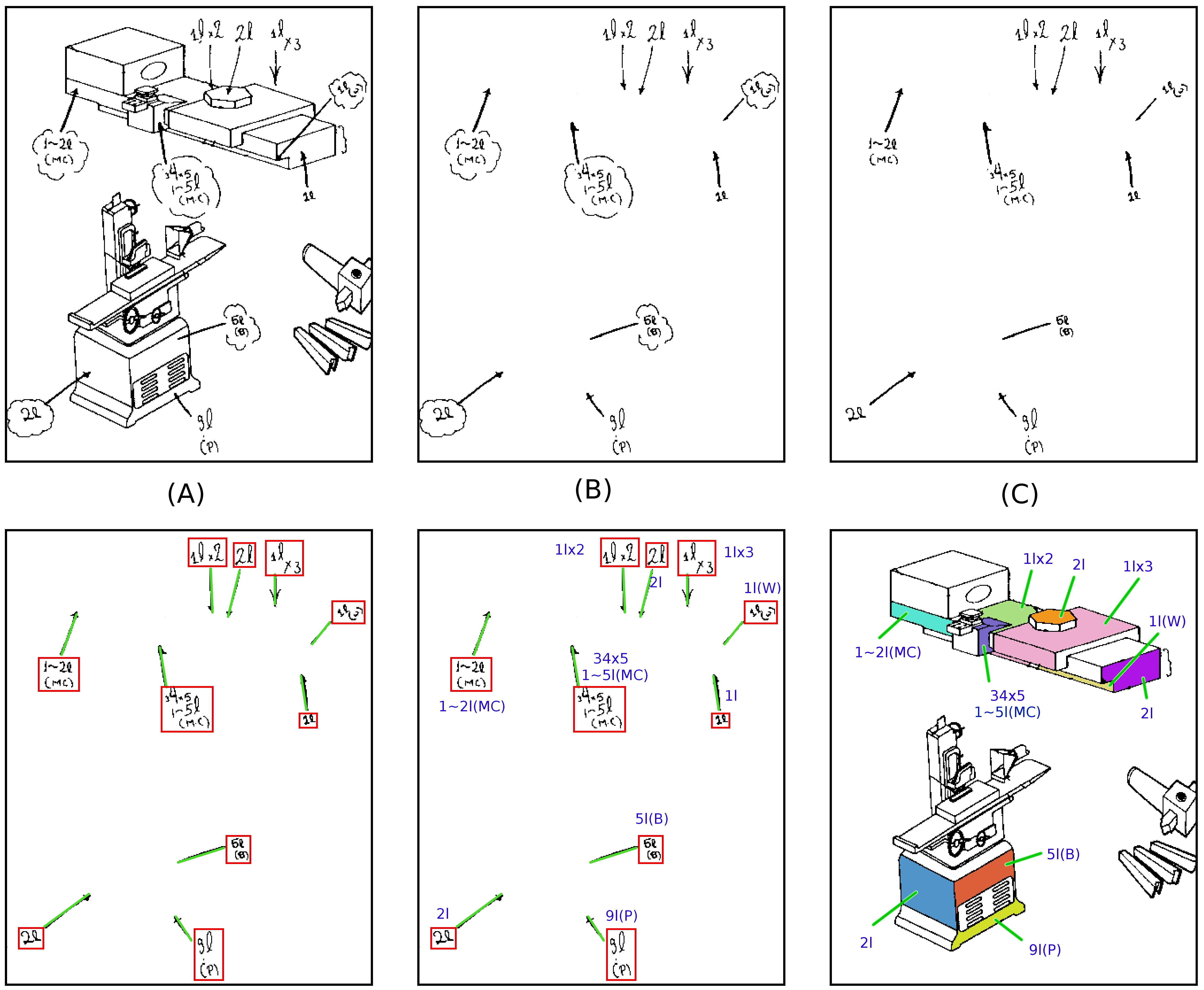}
\caption{Overview of the framework: (A) Original Input Image (B) Template Removal (C) Cloud Removal (D) Text and Arrow Localization (E) Text Reading (F) Text-to-Zone Association}
\label{fig:overview}
\end{figure}

\noindent We propose a novel framework for extracting damage codes, handwritten by a user on an inspection sheet and then associating the same with correponding zones, as shown in Figure \ref{fig:overview}. The major components of our model are described in detail in this section. We first remove the templates and the clouds. Then, we localize the text patches and the connectors. Further, we combine the information on the connectors and text patches for more accurate localization and extraction of the text. This is followed by reading of the damage codes. Finally, we associate the damage codes with the zones, leaveraging the knowledge about the connectors. This process successfully establishes a one-to-one mapping between the zones and corresponding damage codes.

\subsection{Template Extraction and Removal}

\begin{figure}
\centering
\begin{minipage}{.5\textwidth}
  \centering
  \includegraphics[width=.95\linewidth]{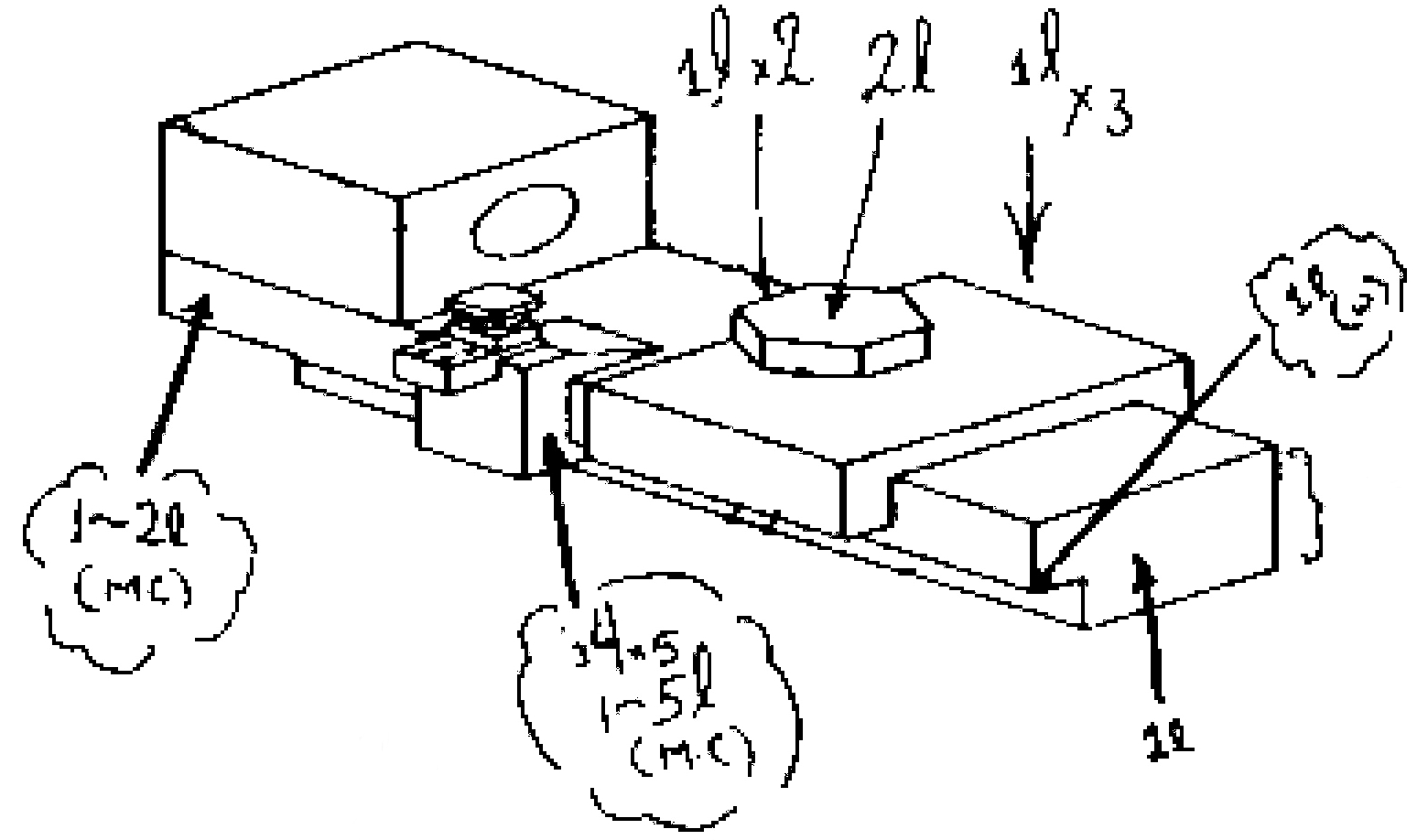}
  \caption{Original Image}
  \label{fig:input_image}
\end{minipage}%
\begin{minipage}{.5\textwidth}
  \centering
  \includegraphics[width=.95\linewidth]{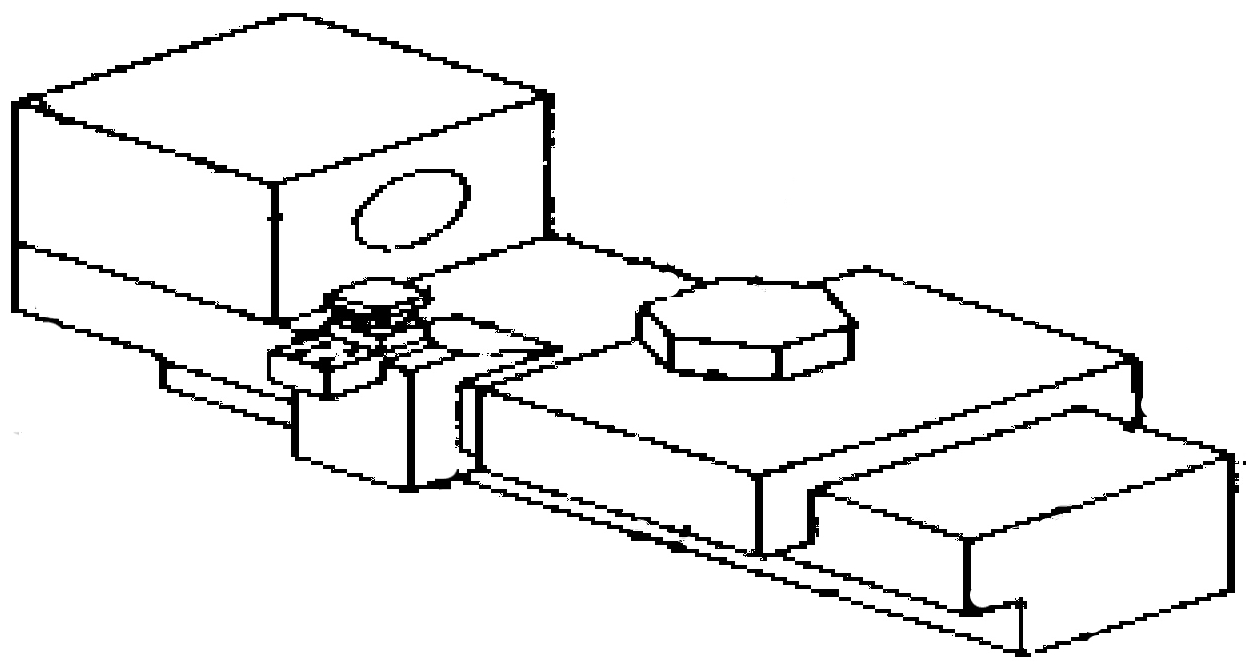}
  \caption{Template in the image}
  \label{fig:temp}
\end{minipage}
\end{figure}

An inspection sheet is essentially composed of a static and a dynamic part. The static part is the 3D orthogonal view of a machine that remains constant over a set of inspection sheets. On the other hand, the dynamic part consists of arrows, clouds and text that is added by the user on top of the static part, as shown in Figure \ref{fig:input_image}. Our goal is to find specific components of the dynamic part and to identify relationships among those parts. We have found that at times static part interferes with the detection of the dynamic part and therefore, as a first step, we remove the static part from the input images.\\ 
  
\noindent\textbf{Template Extraction} : 
Having established the presence of static and dynamic parts in a particular set of sheets, we automate the process of extracting the templates in the sheet. The process involves invertion of the images followed by depthwise averaging and a final step of adaptive thresholding. This generates an image containing just the template. We have noticed that though there are multiple sheets with similar templates, the relative start point of each template is not consistent among the sheets. Hence there is a need to find the individual templates and localize them in the input image. To this end, we find contours on the depth averaged image and then arrange all the detected contours in a tree structure with the page being the root node. In such an arrangement, all nodes at depth $1$ are the templates.\\ 
\\

\begin{figure}
 \centering
 \includegraphics[scale={0.15}]{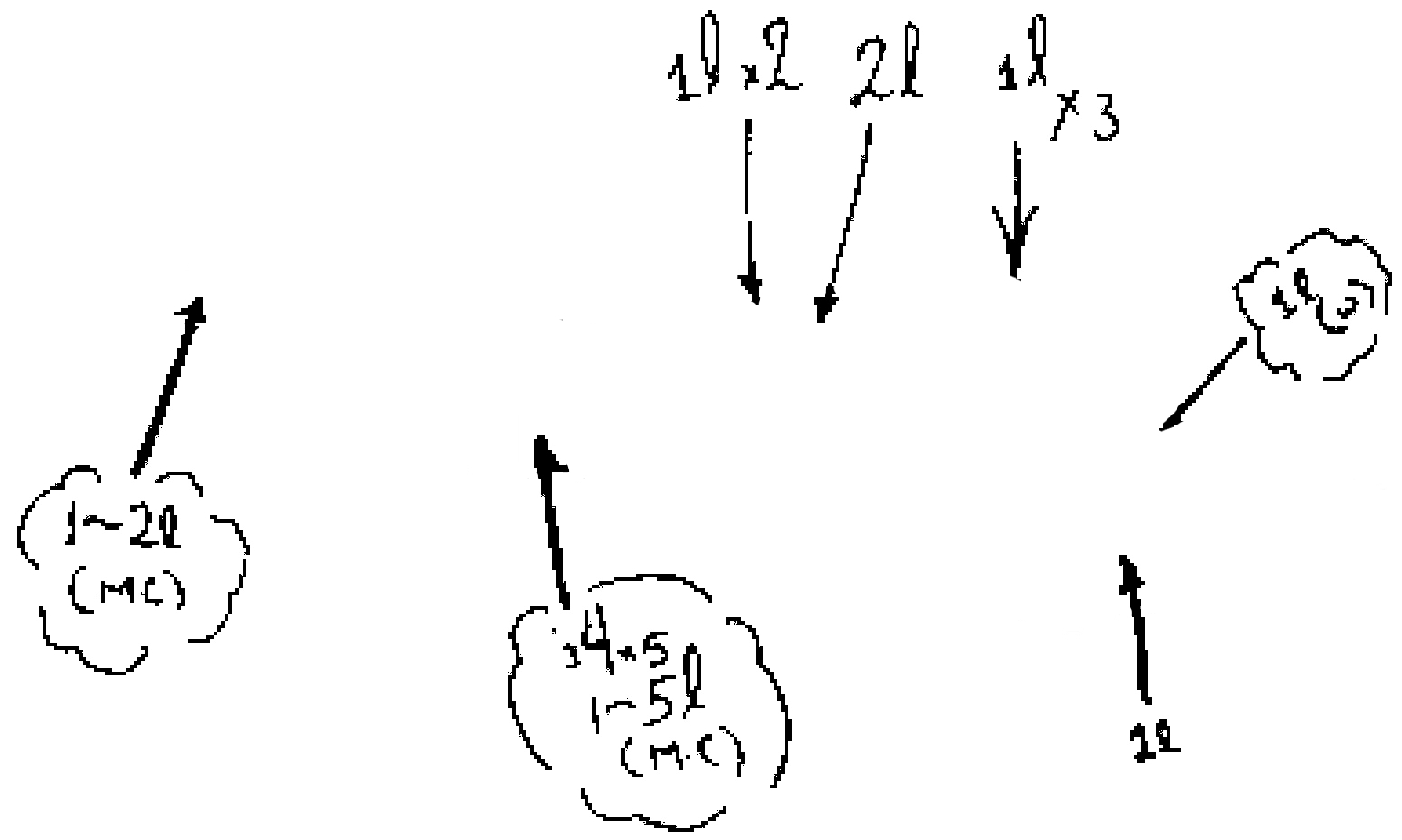}
 \caption{Image with template removed}
\label{fig:tempRem}
\end{figure}
 
\noindent\textbf{Template Localization} : Now that we have the templates, we use \textit{Normalized Cross Correlation} \cite{Yoo:2009:FNC:2740309.2740484} to match the templates with input sheets. This gives us the correlation at each point in the image. By taking the point exhibiting maximum correlation, we can find the location of the template present in the image.

\[
	R(x,y) = \frac{\sum_{x',y'}(T(x',y')*I(x+x',y+y'))}{\sqrt{\sum_{x',y'}T(x',y')^2*\sum_{x',y'}I(x+x',y+y')^2}}
\]
\newline
\noindent\textbf{Template Subtraction} :
To remove the template that was localized in the previous step we use the operator Not(T(i, j)) and R(i, j) on two images T and R, where T is the template image and R is the input image. The resulting image after template subtraction is shown in Figure \ref{fig:tempRem}. 

\subsection{Dialogue cloud segmentation and Removal}
\begin{figure}
 \centering
 \includegraphics[scale={0.35}]{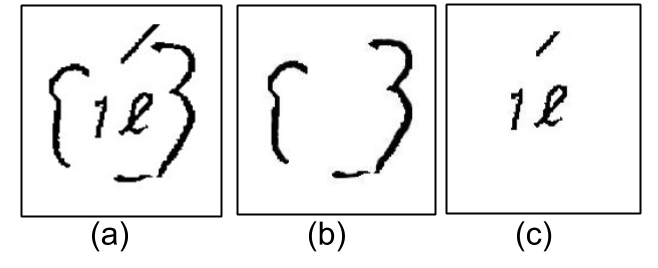}
 \caption{Cloud segmentation a. input image b. mask generated by segnet c. mask removed}
 \label{fig:cloudSeg}
\end{figure}

Dialogue cloud contains the text/comment in documents as shown in Figure \ref{fig:cloudSeg}. They are present sporadically in the inspection sheet and interfere with the detection of dynamic parts like connectors and text. We have used the encoder-decoder based \textit{SegNet} \cite{badrinarayanan2017segnet} architecture for segmenting out dialouge clouds. It was trained on a dataset of $200$ cloud images to distinguish $3$ classes namely background, boundary and cloud. Generally, it was able to learn the structure of the cloud. At times, the segnet would classify a few pixels as background which would lead to introduction of salt and pepper noise around the place where the cloud was present, but we address this issue while text reading by performing median filtering.

\subsection{Connector Detection and Classification}\label{sec:conDec}

Connectors established a one-to-one relationship between text and its corresponding zone. They sometimes manifest as arrows with a prominent head but often they are just lines or multiple broken pieces of a line, as shown in the image, making the automation process far more complex.
We tackle this problem using two approaches : \\ 
\\	1. CNN to detect the arrows with prominent heads
\\	2. Detection of Lines \\ 

\noindent\textbf{Arrow Classification} : As the first step we extract all the connected components from the image to send them to our classifier. We train the \textit{Convolutional Neural Network} (CNN) on $2$ classes which are Arrow and background. We modified the architecture of \textit{Zeiler-Fergus} Network (ZF) \cite{zeiler2014visualizing} and show that our network outperforms ZF network in the task of arrow classification by a considerable margin. We trained the classifier to learn the features of the connectors which have a prominent arrow like head. We observed that including the connectors which do not have a prominent head (i.e they are just a line) confuses all CNN models and the precision falls dramatically. To detect arrows in the input image, we feed each connected component found after the template removal to the CNN classifier. All the connected components that are arrows and have a prominent head are classified as such, subsequently, we use the information of the text patches to find out the head and tail point of the arrow. \\ 

\begin{figure}[!b]

 \centering
 \includegraphics[scale={0.1}]{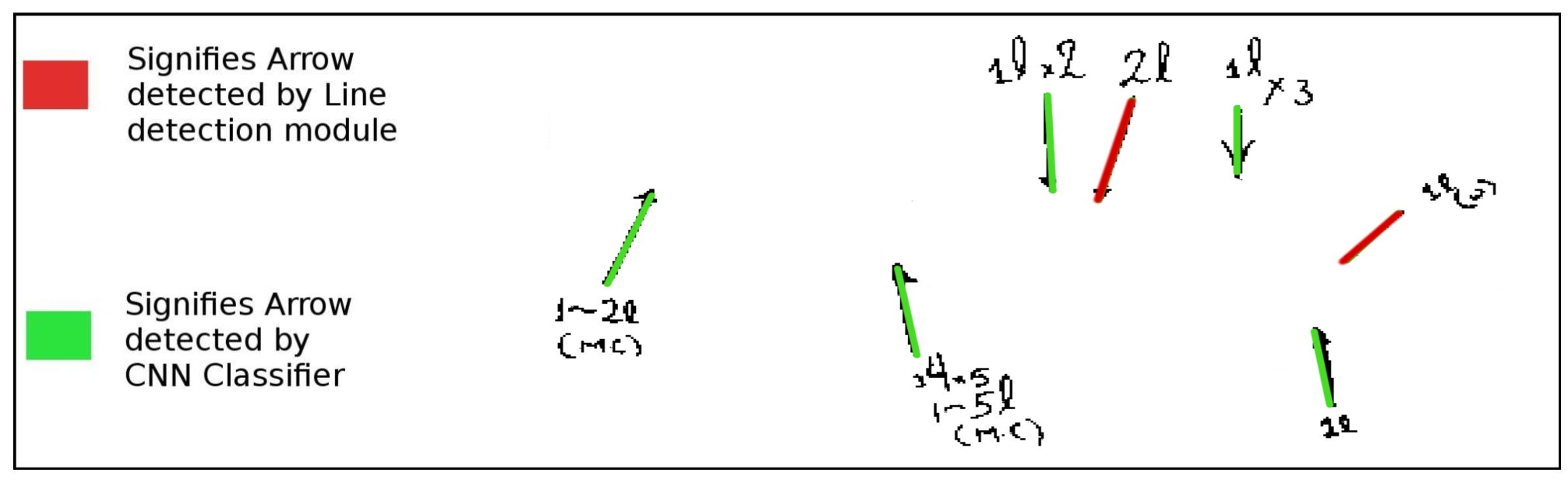}
 \caption{Arrow detection}
 \label{fig:detected_arrow}
\end{figure}

\noindent\textbf{Line Detection} : Most of the arrows having a prominent head would be, at this point, detected by the arrow CNN.  Here, we describe the process of detecting arrows that have been drawn as a line (i.e. without a prominent head ). To this end, we use a three-step approach. The first step involves detection of various lines that were present in the input image after the removal of templates through hough lines. This is followed by line merging and line filtering where we filter the lines based on the association with the text patch. The filtering step is required because a lot of noise would also be detected as lines which can be filtered leveraging the knowledge gained after text patch detection and association. We further elaborate on the filtering step in Section \ref{sec:conFilt}.\\

\noindent\textbf{Line Merging} : As can be seen in Figure \ref{fig:tempRem} that after template removal a lot of arrows are broken into segments and hence for each segment a seperate line would be detected. As a result there would be multiple lines for a single arrow. Therefore we merge the lines if they have the same slope and the euclidean distance between them is within $50$ px. The resulting image after arrow classification and line detection and merging is shown in Figure \ref{fig:detected_arrow}.

\subsection{Text Patch Detection}

\begin{figure}
 
 \includegraphics[scale=0.11]{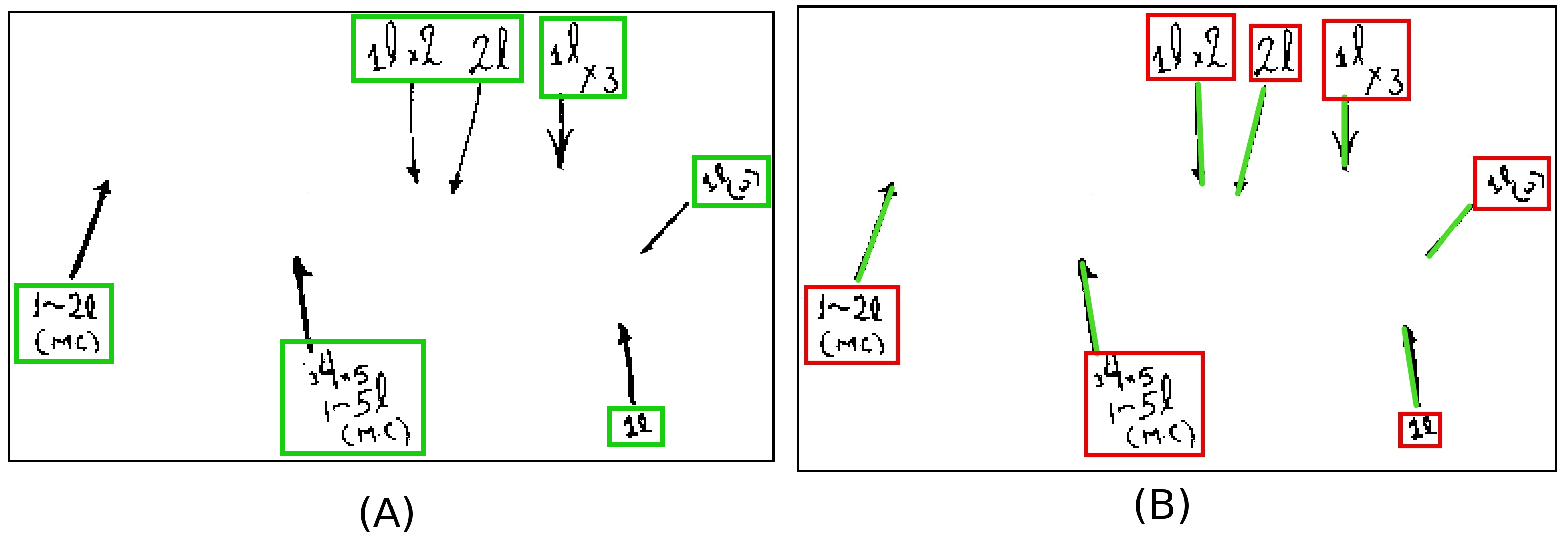}
 \centering
 \caption{(A) Text boxes detected by CTPN (B) Final text boxes after associaton with connectors}
 \label{fig:ctpnOut}
\end{figure}

The text patches in the input image is usually present in the vicinity of a template. To detect these text patches, we employ \textit{Connectionist Text Proposal Network} (CTPN) \cite{tian2016detecting} which has proven to be quite effective in localizing text lines in scene images. With a bit of fine-tuning the same model was able to locate text boxes in the input images. Initially, we trained the CTPN on full size images but it failed to produce desired results. It captured multiple text patches, that occur colinearly, in a single box. This anomaly resulted from the low visual resolution of the individual text patches when looked at from a global context which is the entire image. The network simply captured any relevant text as a single item if they are horizontally close. As a resolution of the same, we sample 480x360 px windows from the input image with overlap. These windows offer better visual seperation between two colinear text patches, resulting in superior localization. 
Nevertheless, not all text boxes that contained more than one text patch can not be eliminated by the same, as shown in Figure \ref{fig:ctpnOut}(A).

\subsection{Connector Filtering and Text Patch Association}\label{sec:conFilt}

For complete resolution of the problem discussed in last section, we use the available information from the detected arrows, as each text patch must have a corresponding arrow tail pointing to it. We associate each arrow to one of the bounding boxes by extrapolating the arrow tails. Once all the detected arrows are associated to a bounding box, we cluster the text patches present, with the number of clusters being equal to the number of arrows associated to that bounding box. This means that if there exists a bounding box that has two or more arrows associated to it, we will obtain the same number of text patches as the number of arrows. We use K-means clustering for this purpose, where K is the number of arrows associated to a CTPN bounding box. This ensures that there will always be only one text patch associated to a single arrow, as shown in Figure \ref{fig:ctpnOut}(B). Once we have the bounding boxes of the text patches, we extract them and send them to the reading pipeline.

\subsection{Text Reading}

\begin{figure}
 \centering
 \includegraphics[scale={0.30}]{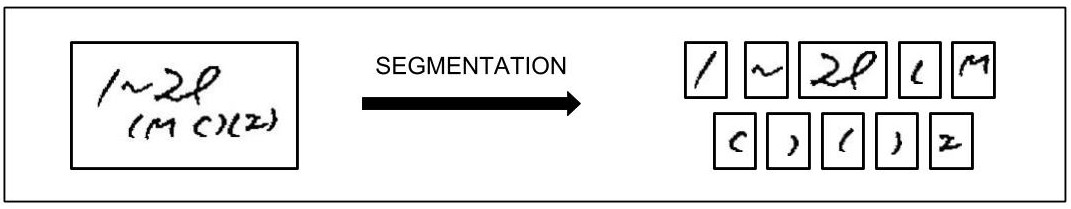}
 \caption{Segmentation output of text patch}
 \label{fig:seg}
\end{figure}

This section describes the text reading component of our model. Input to this system is a set of text patches extracted from the inspection sheets. Each patch contains handwritten alpha-numeric codes corresponding to a particular kind of physical damage. Major challenges arise from the fact that these damage codes are not always structured horizontally in a straight line but consist of multiple lines with non-uniform alignments, depending on the space available to write on the inspection sheets, as shown in Figure~\ref{fig:seg}. Moreover, the orientation of the characters in these codes are often irregular making the task of reading them even more difficult.\\
Due to these irregularities, it was difficult to read an entire text sequence as a whole. Instead, we designed our model to recognize one character at a time and then arrange them in proper order to generate the final sequence. The model consists of a segmentation module that generates a set of symbols from the parent image in no particular order, followed by a ranking mechanism to arrange them in standard human readable form. We then employ two deep neural networks to recognize the characters in the sequence. The final component is a correction module that exploits the underlying syntax of the codes to rectify any character level mistake in the sequence.\\

\noindent\textbf{Segmentation} of individual characters in the image patch is performed using \textit{Connected Component Analysis}( CCA ). As CCA uses a region growing approach, it can only segment out characters that neither overlap nor have any boundary pixels in common. So, the CCA output may have one or more than one characters in a segment. In our experiments, we found that the segments had a maximum of two characters in them.\\

\begin{figure}
 \centering
 \includegraphics[scale={0.28}]{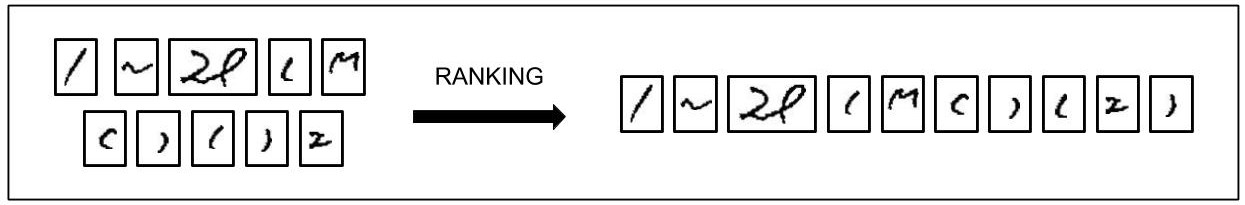}
 \caption{Segmented output of text patch ranked in human readable format}
 \label{fig:ccRank}
\end{figure}

\noindent\textbf{Ranking} of segmented characters is described in Algorithm \ref{alg:algorithm1}. It takes a list of unordered segments and returns another that has the characters arranged in a human readable form i.e. left-to-right \& top-to-bottom, as shown in Figure \ref{fig:ccRank}. \\

\begin{figure}[!h]
 \centering
 \includegraphics[scale={0.36}]{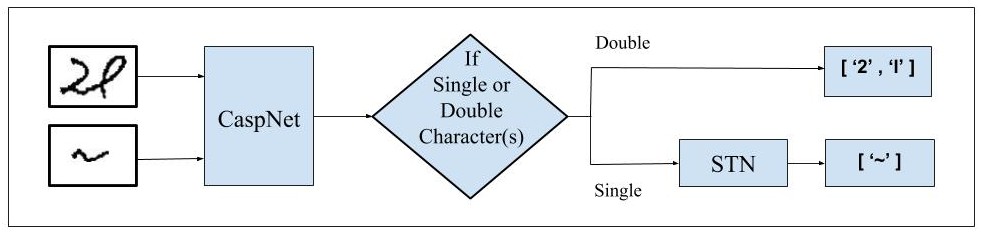}
 \caption{Capsnet and STN}
 \label{fig:capsSTN}
\end{figure}

\noindent\textbf{Character Recognition} is implemented as a two-step process. First step is to determine whether a segment contains one or two characters. Towards this end, we use \textit{Capsule Network} ( CapsNet ) \cite{sabour2017dynamic} which performs remarkably well in classifying multiple characters with considerable overlap. We modified the standard formulation of CapsNet by introducing a new output class, \textit{None} representing the absence of any character in the image. Therefore, in case there is only a single character present in the segment, CapsNet predicts \textit{None} as one of the two classes. In spite of being a powerful classification model, the performance of CapsNet on the test data was limited. This necessitated the second step in which we use a 
\textit{Spatial Transformer Network} (STN) \cite{jaderberg2015spatial} to recognize single character segments. STN consists of a differentiable module that can be inserted anywhere in CNN architecture to increase its geometric invariance. As a result, STN proved to be more effective in addressing randomness in the spatial orientation of characters in the images, thereby boosting the recognition performance. Finally, in case of segments that had two characters, we take the CapsNet predictions as the output as STN cannot classify overlapping characters. This scheme is described in Figure \ref{fig:capsSTN}.\\

\begin{figure}[!h]
 \centering
 \includegraphics[scale={0.30}]{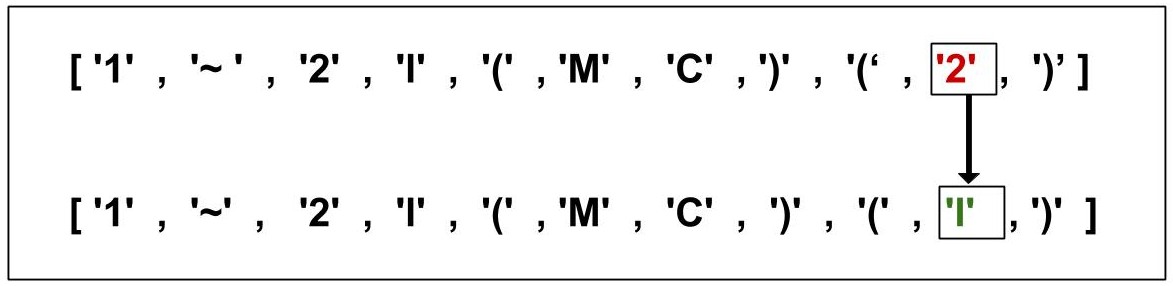}
 \caption{Correction made in the CapsNet and STN predictions based on the grammer of the defect codes}
 \label{fig:correc}
\end{figure}

\noindent\textbf{Correction} module incorporates domain knowledge to augment neural network predictions. It has two parts. First, a \emph{rule-based system} that uses the grammar of the damage codes to rectify predictions of the networks. For example, as per the grammar, an upper case "B" can only be present between a pair of parenthesis, i.e. "(B)". If the networks predict "1B)", then our correction module would correct this part of the sequence by replacing the "1" by a "(". 
On top of it is an \emph{edit-distance based} module which finds the closest sequence to the predicted damage sequence from an exhaustive list of possible damage codes. An example is shown in Figure \ref{fig:correc}.
 
\begin{algorithm} 
\caption{Ranking} \label{alg:algorithm1}
\begin{algorithmic}[1]
\Require $seg\_list[s_1,s_2,s_3,...,s_n]$: Segment $s_i = [  (x_1:x_{m_i}),(y_1:y_{m_i})]$, a collection of pixel coordinates.


        \Procedure{Ranking}{$seg\_list$}
            \State  using $s_i$, calculate extremes and centroid of segmented images, $y_{i\_top},y_{i\_bottom},x_{i\_left},x_{i\_right},y_{i\_cen}$ and the $x_{i\_cen}$ 
            \State sort seg\_list in ascending order of $y_{i\_cen}$
             parts
        \State $line\_bottom \gets y_{i\_top}$ ; $line\_begin\_Idx \gets 0$

            \For{i in (1,len(seg\_list))}
                \If{$y_{i\_top}$ $>$ overlap\_thresh}\State
                {sort previous line based on $x_{i\_left}$ ; $line\_begin\_Idx \gets i$}

                \EndIf
              \State line\_begin\_Idx = max($y_{i\_top}$ , line\_bottom)
            \EndFor
            \State sort last line based on $x_{i\_left}$
            \State \Return $seg\_list$
        \EndProcedure
    \end{algorithmic}
\end{algorithm}

\subsection{Zone mapping}
\begin{figure}[!h]
 \centering
 \includegraphics[scale={0.50}]{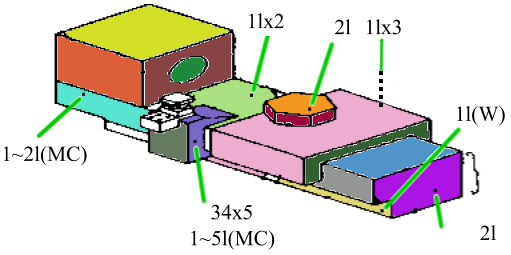}
 \caption{In this example , different zones of a subassembly are shown with different colors and the rays are shown as dotted red lines extending from the arrow head}
 \label{fig:zone}
\end{figure}

After getting the location of the arrows and associating them with the corresponding text, we now have to map the damage codes to the zone. A sample machine part with different zones are shown in Figure \ref{fig:zone}. Here, arrows are used to describe this relationship as the head of an arrow points to the relevent zone and the tail of the same points to the text patch that is to be associated with the corresponding zone. We have already done the latter in the previous section and now we are left with the task of finding out the relevent zone to which the arrow is pointing. We observed that this problem can be  easily solved by ray casting algorithm \cite{shimrat1962algorithm}. If we extend a ray from the head of the arrow, the zone that it intersects first is the relevent zone and can be mapped to the text patch.

\noindent We summerize the proposed method in the following flow diagram : 

\begin{figure}[!h]
\includegraphics[scale={0.142}]{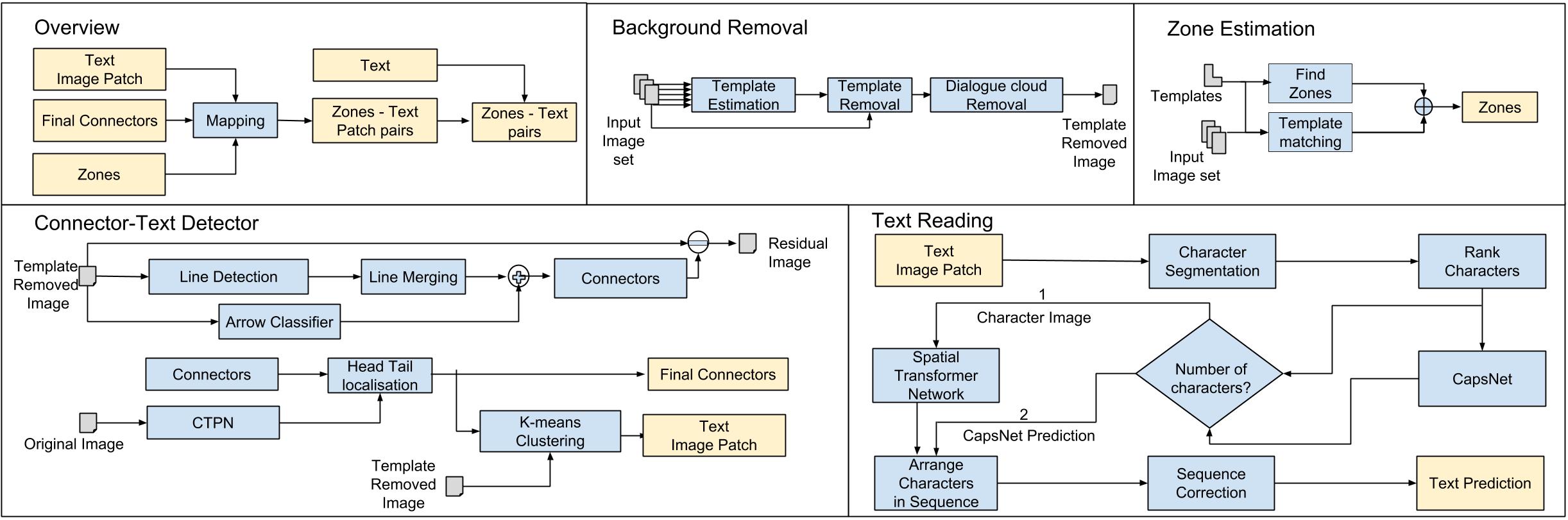}
\caption{Flow diagram of the proposed framework. Each component represents a set of operations that extract a particular information from the input images.}
\label{fig:propMet}
\end{figure}


\section{Experiments}
\label{sec:exp}

\subsection{Implementation Details}

We have a confidential dataset provided by a firm. It has $72$ different kinds of machine structures distributed across $10$ sets of images.
There were $50$ equally distributed images for testing. This implies that a particular set has same machine line diagrams forming the static background. For training purpose, a separate set of $450$ images are kept with same distribution of background machine line diagram sets.
All the sheets are in JPEG format with resolution of $3500 \times 2400$ sq. px. They have been converted into inverted binarized version where the foreground is white and background is black. The conversion is done by Otsu's binarisation.\\ 

\noindent\textbf{Dialogue Cloud Segmentation}:
For this process, we have used the SegNet \cite{badrinarayanan2017segnet} model to train on 200 images. Two classes are cloud pixels and background. As there is an imbalance, the classes are weighted by 8.72 for the foreground and 0.13 for the background.\\

\noindent\textbf{Arrow Classifier}:
The classifier is inspired from \cite{seddati2015deepsketch}. It includes 6 convolution layers and $2$ fully connected layer with ReLU \cite{krizhevsky2012imagenet} activation. Max pool and dropout (with $0.5$ probability) were used for regularization. We set the learning rate of $0.001$ and used the Adam \cite{kingma2014adam} optimizer with cross entropy loss to train it on $800$ images with equal number of images per class. We initialized the network using Xavier initializer \cite{glorot2010understanding} and trained the model till best validation accuracy achieved after $50$ epochs. We used \textit{Batch Normalization} \cite{ioffe2015batch} with every convolution layer so as to make the network converge faster. The network is $99.7$\% accurate on a balanced test set of $400$ images. The input images are resized to  ($128 \times 128$) with padding such that the aspect ratio of the images is undisturbed.\\

\begin{figure}[!h]
 \centering
   \includegraphics[width=0.95\linewidth]{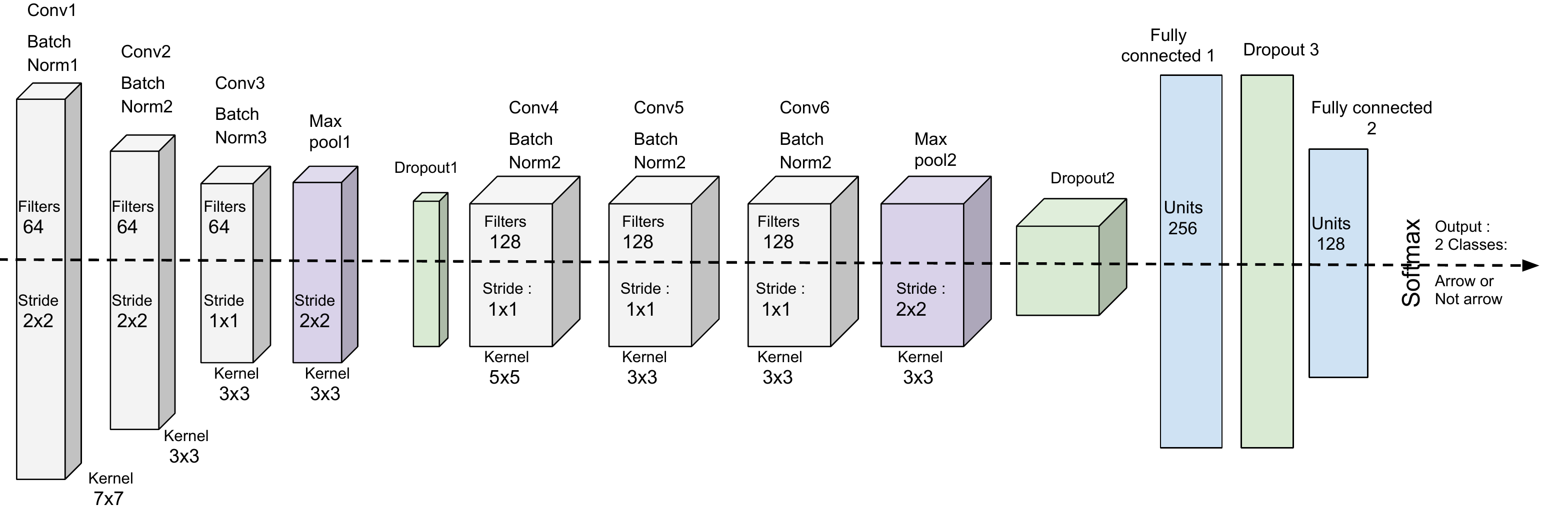}
 \caption{Architecture of CNN used for the arrow classification}
\end{figure}

\noindent\textbf{Capsule Network}:
We have used this network as it has proven to be effective in classifying overlapping characters on the MNIST \cite{deng2012mnist} dataset. We set the learning rate to 0.0005 and use the Adam Optimizer to train our model on all the single characters as well as on all the possible pairs of characters touching each other.\\

\noindent\textbf{Spatial Transformer Network (STN)} 
These are convolutional neural networks, containing one or several Spatial Transformer Modules. These modules try to make the network spatially invariant to its input data, in a computationally efficient manner, leading to more accurate object classification results. We have taken the architecture from \cite{jaderberg2015spatial}. All the input images are padded and resized to $32 \times 32$ so that they do not loose their original aspect ratio. We trained this network on images of all the 31 characters. \\

\subsection{Results}

We present results for individual components as well as the overall performance of the model.
 


\begin{table}[]
\parbox{.45\linewidth}{
\centering
    \caption{Accuracy of Individual Components for Text Extraction and Association}
    \label{tab:results-connector}
    \begin{tabular}{|l|c|} 

\hline
\textbf{Component} & \textbf{Accuracy}\\
\hline
Connector Detection & $89.7\%$  \\
\hline
CTPN &  $91.6\%$ \\
\hline
Patch Association & $95.1\%$  \\
\hline
Clustering & $95.6\%$  \\
\hline
Zone mapping & $96.4\%$  \\
\hline
    \end{tabular}
}
\hfill
\parbox{.45\linewidth}{
\centering
    \caption{Accuracy of Individual Components for Text Reading}
    \label{tab:reading}
    \begin{tabular}{|l|c|c|}
\hline
\textbf{Component} & \textbf{Accuracy}\\
\hline
CCA & $97.54\%$ \\
\hline
Ranking & $98.08\%$ \\
\hline
CapsNet(Overlap) & $66.11\%$ \\
\hline
CapsNet(Non-Overlap) & $89.59\%$ \\
\hline
STN & $95.06\%$ \\
\hline
Sequence Reading& $94.63\%$ \\
\hline 
    \end{tabular}
}
\end{table}

\begin{table}[h]
\caption{Cumulative accuracy for the complete framework }
\centering{
\begin{tabular}{| l | c | c |}

\hline
\textbf{Component} & \textbf{Individual Accuracy} & \textbf{Cumulative Accuracy}\\
\hline
Text Association & $87.1\%$ & $87.1\%$ \\
\hline
Text Reading & $94.63\%$ & $82.3\%$ \\
\hline
\end{tabular}
}\\
\label{tab:results-end2end}
\end{table}

\noindent The results of Connector Detection is shown in Table \ref{tab:results-connector}. A total of 385 arrows were correctly localized out of 429 arrows present. The detection was performed on the sheets where the templates were removed. Majority of the false negatives occured as a result of probabilistic hough lines missing the entire line or most of the line, resulting in its removal during the arrow filtering stage.

\noindent The result of text patch detection using CTPN is shown in Table \ref{tab:results-connector}. It detected 392 text patches out of a total of 429 text patches correctly. It missed a few patches entirely and it resulted in a few false negatives in which it was generating a bounding box enclosing more than a single text patch inside it.

\noindent Out of the 392 text patches that the CTPN detected, 374 were correctly associated with the correct arrow, giving us the Patch Association accuracy as shown in Table \ref{tab:results-connector}.

\noindent And for the boxes which were associated with multiple arrows(false negative of CTPN enclosing more than a single text patch), we applied k-means clustering on the connented components present inside the CTPN boxes. It  resulted in clusters of connented components belonging to the same text patch. Out of 23 such text patches which asked for clustering, k-menas clustering 22 of them correctly yielding an overall accuracy of 95.6\% as shown in Table \ref{tab:results-connector}


\noindent We present the results of the text reading module in Table \ref{tab:reading}. We performed our experiments on 349 image patches. The accuracy of the CCA is calculated as the percentage of correct characters outputs in the total number of outputs. Ranking accuracy is calculated as a percentage of correct rankings done by the total number of images patches. The performance of the capsule network has been measured for two tasks (mentioned in the Table \ref{tab:reading} above), one being the recognition of the overlapping characters and second, character level recognition in cases of non-overlapping characters. And at last the STN accuracy shows the character level accuracy which is better than the character level accuracy of the Capsule Network, justifying the reason why STN was used in the first place. Now the sequence level recognition's accuracy can be measured by measuring the ground-truth as well as the final predictions of the networks passing through both the correction modules, which is shown in the Table \ref{tab:reading}. The way we consider a prediction correct is if and only if all the characters in the predicted string matches with the ground-truth in the correct order.\\
The cumulative accuracy of the framework is provided in Table \ref{tab:results-end2end}.

\section{Conclusion}
\label{sec:conc}

The proposed framework has given a detection accuracy of 87.1\% for detection and
94.63\% for reading. It manages to achieve high accuracy and is robust to different types of noise in arrow / cloud / text detection and character recognition. While it may be possible to train a deep system or model to learn this task in an end to end fashion given a very large set of cleanly annotated documents, but with the limited data at our disposal, incorporation of domain information was mandatory.
As the entire pipeline is dedicated to a given layout, we plan
to formulate an approach that is customizable with different layout types in future.


%
%
\bibliographystyle{splncs04}
\bibliography{egbib}
\end{document}